%% file: main.tex
\documentclass[11pt]{article}

\usepackage[final]{acl}
\usepackage{amssymb}
\usepackage{multirow}

\usepackage{times}
\usepackage{latexsym}
\usepackage{amsmath}

\usepackage{longtable}

\usepackage{fontspec}
\usepackage{xeCJK}

\usepackage[T1]{fontenc}

\usepackage{float}

\usepackage{inconsolata}

\usepackage{graphicx}

\usepackage[capitalise]{cleveref}
\usepackage{url}
\usepackage{booktabs}
\usepackage{bbding}
\usepackage{makecell}

\usepackage{algpseudocode}

\usepackage{pifont} 
\usepackage{amsfonts}\usepackage{mathrsfs}\usepackage{tabularx}
\usepackage{colortbl}
\usepackage{lscape}
\usepackage{wrapfig}
\usepackage{subfig} 

\usepackage{amsmath}
\usepackage[linesnumbered,ruled]{algorithm2e}

\usepackage{mathptmx}
\usepackage[english]{babel}
\usepackage{graphicx}
\usepackage[makeroom]{cancel}
\usepackage{hyperref}

\usepackage[protrusion=false,expansion=false]{microtype}

\usepackage[T1]{fontenc}

\usepackage{colortbl}
\usepackage{booktabs}

\usepackage{tcolorbox}
\tcbuselibrary{breakable}

\usepackage{xcolor}

\definecolor{mc-color}{RGB}{245, 250, 255}
\definecolor{gen-color}{RGB}{247,254,247}

\definecolor{model-color}{RGB}{250, 250, 250}

\definecolor{dark-grey}{RGB}{245,245,245}

\definecolor{white}{RGB}{255,255,255}

\definecolor{mycustomgreen}{RGB}{178,234,3}     
\definecolor{mycustompink}{RGB}{255,225,229} 
\definecolor{mycustomblue}{RGB}{198,232,255}

\title{SenseJudge: Human-Centric Preference-Driven Judgment Framework}

\author{
Rui Li$^1$\thanks{Equal contribution. This work was conducted during an internship at StepFun AI from August 2024 to April 2025.}, Junfeng Liu$^{2,3 *}$, Xiangwen Kong$^2$, Linhai Xu$^3$\thanks{Corresponding authors.}, Zhifang Sui$^1$\footnotemark[2] \\
$^1$\normalsize{State Key Laboratory for Multimedia Information Processing, School of Computer Science,}\\ \normalsize{Peking University} \quad
$^2$\normalsize{StepFun} \quad
$^3$\normalsize{Xi'an Jiaotong University} \quad \\
\texttt{o\_l1ru1@outlook.com, 3124336079ljf@stu.xjtu.edu.cn}\\
}

\begin{document}
\maketitle

\begin{abstract}

Using Large Language Models (LLMs) as judges across scenarios such as model-response assessment is becoming an increasingly accepted paradigm. 
However, existing judgment approaches often rely on trained judgers using fixed preference data, which tend to overlook diverse user preferences and struggle to adapt to real-world human-AI dialogue scenarios.
To address these limitations, we propose \textbf{\textit{SenseJudge}}, a customizable judgment framework driven by human preferences, and  \textbf{\textit{SenseBench}}, a diverse and challenging instruction-following benchmark derived from real-world multi-turn interactions.
We evaluate the framework and benchmark on two tasks: 1) \textit{LLMs as personalized judges}, and 2) \textit{model ranking}.
Extensive experiments show that the SenseJudge framework outperforms the compared judgment methods and models in the LLMs-as-personalized-judges task and produces model rankings that align with human judgments.
Additionally, we conducted analyses on position bias and consistency, alongside ablation studies, which support the robustness of SenseJudge.
Our data and code are publicly available in the \href{https://github.com/qiongrenpiqida/SenseJudge}{GitHub repository}.

\end{abstract}

\section{Introduction}
\input{section/1_introduction}

\section{Related Work}
\input{section/5_relatedwork}

\section{Method}
\input{section/3_method}

\section{Experiments}
\input{section/4_experiments}

\section{Conclusion}
In this work, we introduced SenseJudge, a novel framework leveraging explicit human preferences for customizable and personalized LLM judgment, and SenseBench, a challenging benchmark mirroring real-world human-AI interactions. Extensive experiments in personalized judging and model ranking demonstrate that SenseJudge improves over existing methods and performs strongly on RewardBench, highlighting its potential for more accurate and user-centric LLM evaluation.

\section*{Acknowledgments}
This paper is supported by NSFC project 62476009.

\section*{Limitations}

The LLM-as-a-Personalized-Judge task focused on simulating a limited number of explicit human preferences. Due to the high cost of annotation, we employed three annotators, each labeling the same 1000 comparison data points, rather than expanding to a broader scope. In our experiments, SenseJudge achieved higher accuracy than the compared baselines in simulating the preferences of all three individuals.

\section*{Ethics Statement}

This paper introduces SenseJudge, a framework for personalized and customizable judgment of LLM responses, driven by explicit human preferences. We believe that SenseJudge offers a valuable tool for advancing the understanding and evaluation of LLMs in a more user-centric manner. However, responsible development and deployment necessitate ongoing attention to potential biases, fairness considerations, and the ethical implications of simulating and utilizing human preferences in automated judgment frameworks.
For the benchmark, we have taken measures to anonymize user data within SenseBench to protect individual privacy. The LLMs used in our experiments are existing models with publicly available access. Our use of these models and datasets aligns with their intended research purposes and licenses. Therefore, we believe that our work complies with ACL ethics requirements.

\bibliography{custom}

\appendix

\section{Appendix}

\input{section/appd}

\end{document}

%% file: section/1_introduction.tex
\definecolor{mychosencolor}{RGB}{255,225,229}
\definecolor{myrejectedcolor}{RGB}{227,255,184}

The evaluation of model responses is a crucial component in the development of Large Language Models (LLMs). Selecting the superior response from a pair generates preference data (query, \colorbox{mychosencolor}{chosen}, \colorbox{myrejectedcolor}{rejected}), which is essential for various post-optimization algorithms~\cite{rafailov2024directpreferenceoptimizationlanguage, ouyang2022traininglanguagemodelsfollow}. Competitive assessments of different LLM versions across various companies facilitate model updates and iterations~\cite{lambert2024rewardbench, li2024dissectinghumanllmpreferences}.

Furthermore, by evaluating a variety of model responses and selecting those that most closely align with individual user preferences, the judgment process can accommodate a broad spectrum of user values, effectively delivering a personalized experience.~\cite{fan2025userdrivenvaluealignmentunderstanding,dong2024llmpersonalizedjudge}.

Previously, these judgments were made through manual human annotation. However, with advancements in the natural language understanding and instruction-following capabilities of Large Language Models (LLMs)~\cite{naveed2024comprehensiveoverviewlargelanguage, zhou2023instructionfollowingevaluationlargelanguage}, LLMs are increasingly being integrated into human workflows for automatic judgment~\cite{zheng2023judging, wang2023pandalm}.

\begin{figure}
    \centering
    \includegraphics[width=1\linewidth]{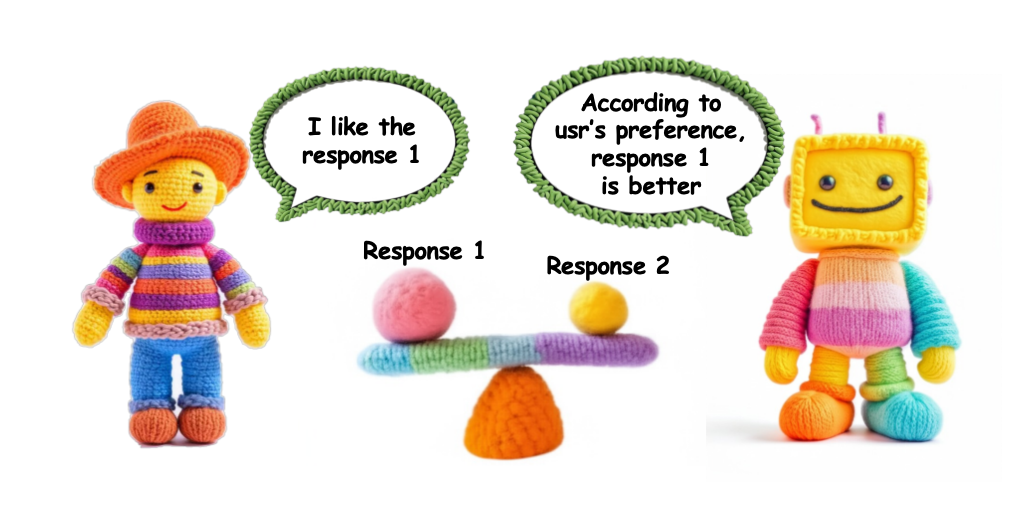}
    \caption{The model judges responses on behalf of human based on individual preferences.}
    \label{fig:personalized-judge}
\end{figure}

However, existing evaluation methods and models, such as PandaLM~\cite{wang2023pandalm}, AutoJ~\cite{li2023generative}, and various reward models, rely on homogeneous standards and values, overlooking the diversity of user preferences. These models also lack the ability to make nuanced choices based on individual user needs. Furthermore, current public judgment benchmarks, including MT-Bench~\cite{zheng2023judging}, AutoJ~\cite{li2023generative}, and RewardBench~\cite{lambert2024rewardbench}, predominantly feature single- or double-turn dialogue designs. This represents a significant departure from real-world human-AI interactions, which often involve multiple rounds of historical dialogue, some relevant and others not, to inform the current request.

To address these limitations, we propose \textbf{\textit{SenseJudge}} and \textbf{\textit{SenseBench}}.
\textbf{\textit{SenseJudge}} is a customizable evaluation framework based on explicit human preferences, enabling personalized judgment of model responses by simulating the preferences of different users and scenarios.
\textbf{\textit{SenseBench}} is a high-quality, challenging dataset derived from realistic human-AI dialogues, covering eight common topics, including math, logic, writing, and role-play.

We conducted extensive experiments to evaluate the effectiveness of the SenseJudge framework in two primary application scenarios based on SenseBench:
1. LLM-as-a-Personalized-Judge, which focuses on selecting responses that align with individual preferences, and
2. Model Ranking, which aims to identify the leading model among different companies or versions (e.g., GPT vs. DeepSeek).
The results demonstrate that SenseJudge outperforms the compared state-of-the-art APIs, training-based evaluators, and reward models in the LLM-as-a-Personalized-Judge task. It also achieves model rankings that align with human judgment.

Additionally, the effectiveness and robustness of SenseJudge are further validated through analyses of factors such as positional bias and consistency, as well as comprehensive ablation studies.

Our main contributions are as follows:

\begin{itemize}
    \item We developed SenseBench, a benchmark designed to align with realistic human-AI interaction scenarios. It features diverse and challenging multi-turn dialogues.

    \item We propose SenseJudge, a judgment framework designed to adapt to diverse human values and preferences, which enables fine-grained and customizable judgment of model responses.

    \item We conducted comprehensive experiments and analysis to evaluate the effectiveness of SenseJudge in the LLMs-as-personalized-judges task and model ranking task. 

    \end{itemize}

%% file: section/5_relatedwork.tex
\label{section_related_work}

\subsection{LLM as a Judge}

LLMs are increasingly recognized as a viable alternative to traditional expert-driven evaluations due to their scalability and adaptability in evaluation tasks. 
Common enhancement methods include designing prompts and training judgers.
Prompt engineering is an effective approach to enhancing the performance of LLMs as judges without updating their model parameters.
Many methods optimize their respective prompt strategies prior to training, such as incorporating references~\cite{zhu2023judgelm} or rules~\cite{li2023generative} into the prompts, and directing models to offer detailed analyses or justifications~\cite{kim2024prometheus}.
Based on this, pairwise or point-wise judgment data can be used to train LLM-based judges by employing supervised fine-tuning~\cite{kim2024prometheus}.
Some researchers train reward models using large-scale preference feedback data; leveraging their ability to learn preferences, these models can also function as judges~\cite{zhong2025comprehensivesurveyrewardmodels}.

\subsection{From Universal to Individualized}
The transition from generalization to personalization represents a historic paradigm in technological development~\cite{you2025framework}. 
With the rapid advancement of the general capabilities of LLMs, there is growing interest in tailoring LLMs to specific user contexts, moving beyond their role as merely general-purpose chatbots~\cite{zhang2024personalizationlargelanguagemodels}.
For instance, ~\citet{wu2024aligningllmsindividualpreferences} proposed to implicitly infer users' personalized preferences through multi-turn conversations and help LLMs to dynamically adapt their behavior and responses to better align with individual user needs.
~\citet{zhu2025personalityalignmentlargelanguage} proposed to use psychological assessment tools, such as the ``Big Five Personality Traits'' and the ``Dark Triad'' to quantify user characteristics and align the models accordingly. 
~\citet{li2025personalizednewsrecommendationmultigranularity} introduced a multi-granularity interest prediction framework which leverages both coarse-grained and fine-grained interest information to supervise model training. 
Additionally, ~\citet{tan2025democratizinglargelanguagemodels} proposed a personalized framework OPPU, which enhances the personalization and generalization capabilities of LLMs by creating dedicated personalized modules for each user through parameter-efficient fine-tuning and non-parametric personalization methods, such as retrieval-augmented and profile-augmented techniques.

%% file: section/3_method.tex
\label{sec:method}

\begin{figure*}
    \centering
    \includegraphics[width=1\linewidth]{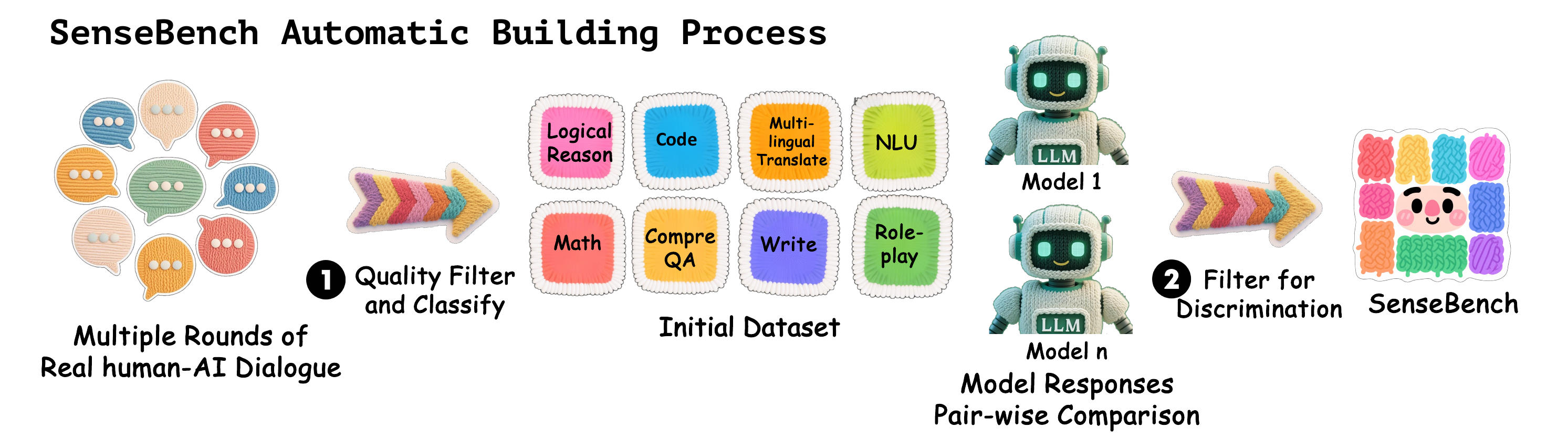}
    \caption{An overview of the data construction pipeline of SenseBench. The pipeline involves 1) Quality-Based Data Filtering and Categorization and 2) Challenge-based Filtering.}
   
    \label{fig:sensebench-pipeline}
\end{figure*}

In this section, we introduce the two key components of our framework: \textbf{SenseBench}, a diverse and high-quality instruction-following benchmark derived from real human-AI interaction dialogues, and \textbf{SenseJudge}, a customizable evaluation method based on human preferences.

\subsection{SenseBench}

\subsubsection{Overview}

Currently, there is a lack of benchmarks that accurately reflect human user experiences and differentiate models based on perceived quality. SenseBench consists of anonymized real user queries sourced from an online LLM service, where the final turn of each dialogue either provides context-dependent instructions or introduces entirely new tasks. This design places high demands on the LLMs' ability to align with human expectations, facilitating effective model differentiation and improving alignment with human-perceived quality.

\subsubsection{Benchmark Construction}

Based on real user data, we employ a scalable automated data selection pipeline to construct a high-quality and diverse benchmark as shown in Figure~\ref{fig:sensebench-pipeline}. The pipeline consists of two main steps: 1) Quality-based Data Filtering and Categorization, and 2) Challenge-based Filtering. The detailed process is as follows:

\paragraph{Quality-Based Data Filtering and Categorization} 
We use Qwen3-14B\cite{yang2025qwen3} to denoise and categorize data from real user dialogue history. Specifically, we first remove redundant, ambiguous, low-quality, or overly simplistic entries and then categorize the remaining data into eight distinct topics: math, logical reasoning, comprehensive question answering, coding, creative writing, multilingual translation, role-playing, and natural language understanding. These categories represent common types of real-world human instructions and provide a comprehensive reflection of the capabilities of LLMs.

\paragraph{Challenge-based Filtering} 

To ensure that the selected prompts are sufficiently challenging, we require that they generate clearly distinguishable responses from different models. We conduct a multi-model evaluation on candidates that have passed an initial round of automated filtering. Specifically, we input each candidate prompt into both relatively stronger models (e.g., Qwen2.5-Max, Qwen3-235B) and relatively weaker models (e.g., Qwen2.5-14B, LLaMA3-8B) to generate responses. This results in a large set of pairwise data in the format of (Query, Strong Model Response, Weak Model Response). We then apply a two-stage filtering process that combines both automatic and manual filtering.

\textbf{\textit{Automatic Filtering.}}
We use GPT-4 to assess the response quality between strong and weak models. Questions where the strong model significantly outperforms the weak model are retained as discriminative and challenging instances. For specific judging instructions, see Appendix~\ref{app:prompt}.
\textbf{\textit{Manual Filtering.}}
After the initial automatic filtering, we performed manual filtering on the data. Specifically, the manual filtering process focuses on two main aspects:
1) Assessing whether the pairwise data filtered by GPT-4 can truly distinguish model capabilities. During this assessment, humans consider visible differences between model responses (e.g., format, length, or level of detail) or other intuitive quality distinctions.
2) Performing partial cleaning on the prompts of the selected data. Since all prompts are sourced from real user data, they may contain formatting issues or improper phrasing, which require manual cleaning to ensure the data is suitable for creating a usable benchmark.

Specific data statistics and category information can be found in Appendix \ref{app:data}.

\subsection{SenseJudge}

\subsubsection{Overview}

SenseBench offers a human-centric environment where we employ multi-model generation followed by pairwise judgment to support two tasks: (1) user-centric preference alignment and (2) model ranking. We introduce SenseJudge, a personalized pairwise judgment approach that demonstrates strong performance in both tasks.
SenseJudge begins by extracting a spectrum of human preferences from a small set of annotated (query, chosen, rejected) pairs. Each preference guides the judge through the same pairwise instances, and the preference set that best matches the human labels is retained as the distilled, generalizable set.
Next, we outline the components of our judgment framework, including input-output format, preference generation, and the selection and application of the preference set.

\subsubsection{Input-Output Format}
For SenseJudge, the input $\mathit{I} = \{q, (r_1, r_2), p\}$ comprises a query $q$, a pair of responses $(r_1, r_2)$, and a preference $p$. The target output $\mathit{O} = \{ \text{judgment}, \text{analysis} \}$ yields a judgment and corresponding analysis, specifically indicating whether $r_1$ or $r_2$ is superior according to the given preference $p$.
Based on previous research~\cite{dong2024llmpersonalizedjudge,zheng2023judgingllmasajudgemtbenchchatbot}, the inclusion of ``ties'' can negatively impact model performance, so we assume that the model can always distinguish between responses based on preference.

\subsubsection{Preference Construction}

We use DeepSeek-R1\cite{guo2025deepseek} to distill implicit human preferences captured in annotated pairwise data into explicit textual preferences. Specifically, we provide a small set of human-annotated pairwise examples to an LLM and prompt it to generate a preference statement explaining why a user might favor one response over the other. Examples of these preferences are provided in Appendix \ref{app:preference}.
One preference is generated for each pairwise sample. These preferences are then incorporated into the data to create a complete personalized evaluation dataset in the format $\mathit{I} = \{q, (r_1, r_2), p\}$, which consists of a query $q$, a pair of responses $(r_1, r_2)$, and a preference $p$. The specific prompt used is detailed in Appendix \ref{app:prompt}.

\subsubsection{Preference Set Selection}

After we obtain the preference set \( P = \{p_1, p_2, \dots, p_m\} \), where \( m \) is the number of sampled examples, we feed each \( p_i \in P \) into the judger and assess its performance on the test set \( T = \{(q_1, r_{11}, r_{12}), \dots, (q_m, r_{m1}, r_{m2})\} \). Note that the set here is the same one we used to extract the preference.
Under each $p_i$, we obtain the judger's selections $C_{i} = \{c_{i1}, c_{i2}, \dots, c_{im}\}$, where $c_{ij} \in \{r_{j1},r_{j2}\}$ indicates that the response $r_{j1}$ or $r_{j2}$ is chosen by the judger.
Comparing the model's selections $C_{\mathcal{M}, i} = \{c_{i1}, c_{i2}, \dots, c_{im}\}$ with the ground-truth human judgments $G = \{g_1, g_2, \dots, g_m\}$, where $g_j \in \{r_{j1},r_{j2}\}$, we obtain a corresponding set of correctness indicators $S_i = \{s_{i1}, s_{i2}, \dots, s_{im}\}$, where $s_{ij} \in \{0,1\}$.
Subsequently, we consider all subsets \(\mathcal{P}_k \subseteq P\) of cardinality \(l\), where the number of preferences \(l\) satisfies \(1 \leq l \leq m\).
For each subset \(\mathcal{P}_k\), we perform a majority vote over its associated correctness indicators \(\{S_{\mathcal{P}_{k1}}, S_{\mathcal{P}_{k2}}, \dots, S_{\mathcal{P}_{kl}}\}\), where \(\mathcal{P}_{ki}\) denotes the \(i\)-th preference in \(\mathcal{P}_k\) and \(S_{\mathcal{P}_{ki}}\) is its corresponding correctness indicator.
This procedure yields an updated set of correctness indicators, denoted by $S_{\mathcal{P}_k}^* = \{s_{\mathcal{P}_{k1}}^*, s_{\mathcal{P}_{k2}}^*, \dots, s_{\mathcal{P}_{kl}}^*\}$, where each $s_{\mathcal{P}_{kj}}^*$ is obtained via a majority-voting scheme applied to the collection $\{s_{k1, j}, s_{k2, j}, \dots, s_{kl, j}\}$.
We then calculate the accuracy of these aggregated correctness indicators $\textbf{Acc}(S_{\mathcal{P}_k}^*) = \frac{1}{m} \sum_{j=1}^{m} \mathbb{I}(s_{\mathcal{P}_{kj}}^* = 1)$, $\mathbb{I}$ is an indicator function. The subset $\mathcal{P}^*_k$ that achieves the highest accuracy $\textbf{Acc}(\mathcal{P}^*_k)$ effectively captures the annotator's underlying preference and can be generalized to unseen examples.

\subsubsection{Preference Set Application}

Given the optimal preference subset $\mathcal{P}^*_k$ obtained from the previous stage, we apply it to an unseen test set $T' = \{(q'_1, r'_{11}, r'_{12}), \dots, (q'_n, r'_{n1}, r'_{n2})\}$, where $n$ denotes the number of test instances. For each preference $p \in \mathcal{P}^*_k$, every test sample $(q'_j, r'_{j1}, r'_{j2})$ is fed into the judger conditioned on $p$, producing a set of model selections $C'_p = \{c'_{p1}, c'_{p2}, \dots, c'_{pn}\}$, where $c'_{pj} \in \{r'_{j1}, r'_{j2}\}$. We then aggregate the outputs across all preferences in $\mathcal{P}^*_k$ by performing a majority vote at the instance level, yielding the final decision
\[
c'^*_{j} = \text{MajorityVote}(\{c'_{p_1 j}, c'_{p_2 j}, \dots, c'_{p_l j}\}),
\]
where $l = |\mathcal{P}^*_k|$. This produces the final aggregated decision set $C'^* = \{c'^*_{1}, c'^*_{2}, \dots, c'^*_{n}\}$, which represents the model’s preference-aligned selections on the test set.

%% file: section/4_experiments.tex
\label{sec:exp}

\begin{table*}[t]
\centering
\newcolumntype{C}{>{\columncolor{mc-color}}c}
\newcolumntype{G}{>{\columncolor{gen-color}}c}
\resizebox{1\textwidth}{!}{
\begin{tabular}{lcccccccc|c}
\toprule
\textbf{Model} & \textbf{Math} & \textbf{Code} & \textbf{Logic} & \textbf{QA} & \textbf{Write} & \textbf{Role} & \textbf{NLU} & \textbf{Trans} & \textbf{Overall} \\
\midrule
\multicolumn{10}{C}{\textbf{Strong Open/Closed-Source Models}} \\
\midrule
\textbf{GPT-4o} & 66.00 & 61.60 & 65.47 & 72.93 & 60.80 & 63.20 & 65.47 & 56.40 & 63.98 \\
\textbf{DeepSeek-V3} & 72.80 & 62.27 & 66.67 & 77.07 & 62.67 & 64.40 & 64.80 & 61.87 & 66.57 \\
\textbf{DeepSeek-R1} & 53.33 & 55.60 & 61.33 & 53.47 & 46.93 & 50.53 & 51.73 & 50.67 & 52.95 \\
\textbf{Qwen-Plus-Latest} & 66.00 & 59.60 & 61.20 & 67.20 & 55.47 & 58.13 & 60.40 & 59.33 & 60.92 \\
\midrule
\multicolumn{10}{>{\cellcolor[rgb]{1,1,0.95}}c}{\textbf{Trained Judger / Reward Models}} \\
\midrule
\textbf{Autoj-13B} & 64.40 & 52.67 & 55.73 & 53.87 & 39.33 & 42.40 & 48.00 & 36.40 & 49.10 \\
\textbf{Skywork-Critic-Llama3.1-70B} & 63.73 & 63.87 & 63.73 & 69.87 & 59.67 & 59.47 & 57.33 & 55.47 & 61.35 \\
\textbf{INF-ORM-Llama3.1-70B} & 67.20 & 63.20 & 64.80 & 69.07 & 60.00 & 54.13 & 58.40 & 59.73 & 62.07 \\
\textbf{QRM-Gemma-2-27B} & 69.60 & 58.93 & 63.47 & 54.13 & 52.53 & 50.40 & 56.53 & 59.73 & 58.17 \\
\textbf{Skywork-Reward-Gemma2-27B} & 70.40 & 61.60 & 66.10 & 74.10 & 64.00 & 60.00 & 62.70 & 58.40 & 64.70 \\

\midrule
\multicolumn{10}{G}{\textbf{SenseJudge}} \\
\midrule
\textbf{Qwen2.5-14B-Instruct} & 60.27 & 60.80 & 59.33 & 70.13 & 61.60 & 62.00 & 64.40 & 57.20 & 61.97 \\
\rowcolor[rgb]{0.96,0.96,0.96} \textbf{ +SenseJudge} & 68.86\scalebox{0.8}{\color[HTML]{32CD32}{(+8.59)}} & 76.27\scalebox{0.8}{\color[HTML]{32CD32}{(+15.47)}} & 73.20\scalebox{0.8}{\color[HTML]{32CD32}{(+13.87)}} & 76.52\scalebox{0.8}{\color[HTML]{32CD32}{(+6.39)}} & 68.00\scalebox{0.8}{\color[HTML]{32CD32}{(+6.40)}} & 67.26\scalebox{0.8}{\color[HTML]{32CD32}{(+5.26)}} & 72.67\scalebox{0.8}{\color[HTML]{32CD32}{(+8.27)}} & 63.07\scalebox{0.8}{\color[HTML]{32CD32}{(+5.87)}} & 70.73\scalebox{0.8}{\color[HTML]{32CD32}{(+8.76)}} \\

\textbf{Qwen2.5-72B-Instruct} & 63.87 & 61.47 & 61.87 & 76.27 & 60.13 & 66.40 & 64.40 & 57.07 & 63.93 \\
\rowcolor[rgb]{0.96,0.96,0.96} \textbf{ +SenseJudge} & 73.47\scalebox{0.8}{\color[HTML]{32CD32}{(+9.60)}} & 68.64\scalebox{0.8}{\color[HTML]{32CD32}{(+7.17)}} & 72.09\scalebox{0.8}{\color[HTML]{32CD32}{(+10.22)}} & 78.67\scalebox{0.8}{\color[HTML]{32CD32}{(+2.40)}} & 65.11\scalebox{0.8}{\color[HTML]{32CD32}{(+4.98)}} & 69.16\scalebox{0.8}{\color[HTML]{32CD32}{(+2.76)}} & 69.60\scalebox{0.8}{\color[HTML]{32CD32}{(+5.20)}} & 73.00\scalebox{0.8}{\color[HTML]{32CD32}{(+15.93)}} & 71.22\scalebox{0.8}{\color[HTML]{32CD32}{(+7.29)}} \\

\textbf{Llama3.1-8B-Instruct} & 62.13 & 58.80 & 59.73 & 72.00 & 58.13 & 58.53 & 60.13 & 53.60 & 60.33 \\
\rowcolor[rgb]{0.96,0.96,0.96} \textbf{ +SenseJudge} & 74.02\scalebox{0.8}{\color[HTML]{32CD32}{(+11.89)}} & 64.90\scalebox{0.8}{\color[HTML]{32CD32}{(+6.10)}} & 73.00\scalebox{0.8}{\color[HTML]{32CD32}{(+13.27)}} & 76.46\scalebox{0.8}{\color[HTML]{32CD32}{(+4.46)}} & 65.96\scalebox{0.8}{\color[HTML]{32CD32}{(+7.83)}} & 64.79\scalebox{0.8}{\color[HTML]{32CD32}{(+6.26)}} & 67.51\scalebox{0.8}{\color[HTML]{32CD32}{(+7.38)}} & 56.93\scalebox{0.8}{\color[HTML]{32CD32}{(+3.33)}} & 67.94\scalebox{0.8}{\color[HTML]{32CD32}{(+7.61)}} \\

\textbf{Qwen3-14B-Instruct} & 71.73 & 64.80 & 69.20 & 78.00 & 61.47 & 67.60 & 66.27 & 56.13 & 66.90 \\
\rowcolor[rgb]{0.96,0.96,0.96} \textbf{ +SenseJudge} & 71.39\scalebox{0.8}{\color[HTML]{32CD32}{(-0.34)}} & 67.24\scalebox{0.8}{\color[HTML]{32CD32}{(+2.44)}} & 72.13\scalebox{0.8}{\color[HTML]{32CD32}{(+2.93)}} & 77.95\scalebox{0.8}{\color[HTML]{32CD32}{(-0.05)}} & 73.73\scalebox{0.8}{\color[HTML]{32CD32}{(+12.26)}} & 71.07\scalebox{0.8}{\color[HTML]{32CD32}{(+3.47)}} & 69.41\scalebox{0.8}{\color[HTML]{32CD32}{(+3.14)}} & 79.07\scalebox{0.8}{\color[HTML]{32CD32}{(+22.94)}} & 72.75\scalebox{0.8}{\color[HTML]{32CD32}{(+5.85)}} \\

\bottomrule
\end{tabular}
}
\caption{Comparative evaluation of various judgment methods and judgers across different categories: Math, Code, Logic, QA (Question Answering), Write, Role (Role-playing), NLU (Natural Language Understanding), Trans (Translation), and Overall their average performance.}
\label{tab:main}
\end{table*}

In this section, we present the performance of SenseJudge in two primary tasks: 1) LLM-as-a-Personalized-Judge and 2) model ranking.

The effectiveness and robustness of SenseJudge are validated through analyses of factors such as positional bias and consistency, as well as comprehensive ablation studies.

\subsection{Experiments Setup}

\subsubsection{Baselines}
\paragraph{LLM+Prompt} 
Direct prompting without preferences across multiple closed-source and open-source models. The closed-source models included GPT-4o and Qwen-Plus-Latest. The open-source models tested were DeepSeek-V3~\cite{deepseekai2025deepseekv3technicalreport}, DeepSeek-R1~\cite{deepseekai2025deepseekr1incentivizingreasoningcapability}, Qwen2.5-14B-Instruct

~\cite{qwen2025qwen25technicalreport}
, Qwen2.5-72B-Instruct

, Llama3.1-8B-Instruct

~\cite{grattafiori2024llama3herdmodels}, and Qwen3-14B-Instruct.

\paragraph{LLM+Training} We compare against publicly available trained judgers such as AutoJ~\cite{li2023generative}

, as well as top-ranking reward models on the RewardBench leaderboard, including the generative model Skywork-Critic-Llama-3.1-70B

, and classifier models including INF-ORM-Llama3.1-70B

, Skywork-Reward-Gemma-2-27B-v0.2

and QRM-Gemma-2-27B

. PandaLM~\cite{wang2023pandalm} and JudgeLM~\cite{zhu2023judgelm} are excluded from the comparison due to token limit constraints.

\subsubsection{Datasets and Metrics}

We use the human-annotated personalized pairwise dataset for the LLM-as-a-Personalized-Judge task, consisting of 1,000 examples (125 per category across 8 categories). We selected 12 widely used or state-of-the-art LLMs (e.g., DeepSeek-R1, Qwen2.5-Max, Grok-3, GPT-3.5, Qwen2.5-7B, Claude-3-Sonnet, Gemini-2.0-Flash-Thinking-Exp, DeepSeek-V3\cite{liu2024deepseek}, GLM-4-Plus, Gemini-1.0-Pro, Moonshot-v1-8k\cite{team2025kimi}, and Minimax-5.5) to generate responses on SenseBench.
From each category, we sample 10 examples for preference generation and subsequent steps. Three annotators with diverse professional backgrounds provide ground-truth labels, which we use to measure accuracy.
For the task of Model Ranking,
we directly used the relative rankings of models on
the human preference leaderboard as the ground truth. We refer to this dataset as the
Ranking Set. We employed specific model comparison examples from the Ranking Set and evaluated the judger's performance by computing the consistency between the rankings produced by the judge and the human-annotated ground-truth rankings.
Furthermore, we also evaluate on the public preference benchmark RewardBench~\cite{lambert2024rewardbench}. 
Response accuracy was used as the evaluation metric. When assessing the correctness of the model's judgment, we employed a strict matching strategy, whereby the target string (e.g., the final decision is response A) must be fully contained within the model's response to be recorded as correct.

\subsubsection{Implementation Details}

We adopt the (query, \colorbox{mychosencolor}{chosen},  \colorbox{myrejectedcolor}{rejected}) format commonly used in pairwise preference data. We test and calculate the overall accuracy by evaluating the chosen response and rejected response in both their original order and their reversed order. We directly employ small-scale LLMs as our judgers without any additional training to balance performance and cost.

For the model ranking task, we randomly sample five preferences from each category within the preference set corresponding to individual annotators (three annotators in total). Each sampled preference is independently applied to the model ranking task using the SenseJudge framework. The final decision for each comparison is obtained via a majority voting scheme over the predictions induced by these preferences.

\begin{figure*}[t]
    \centering
    \includegraphics[width=1\linewidth]{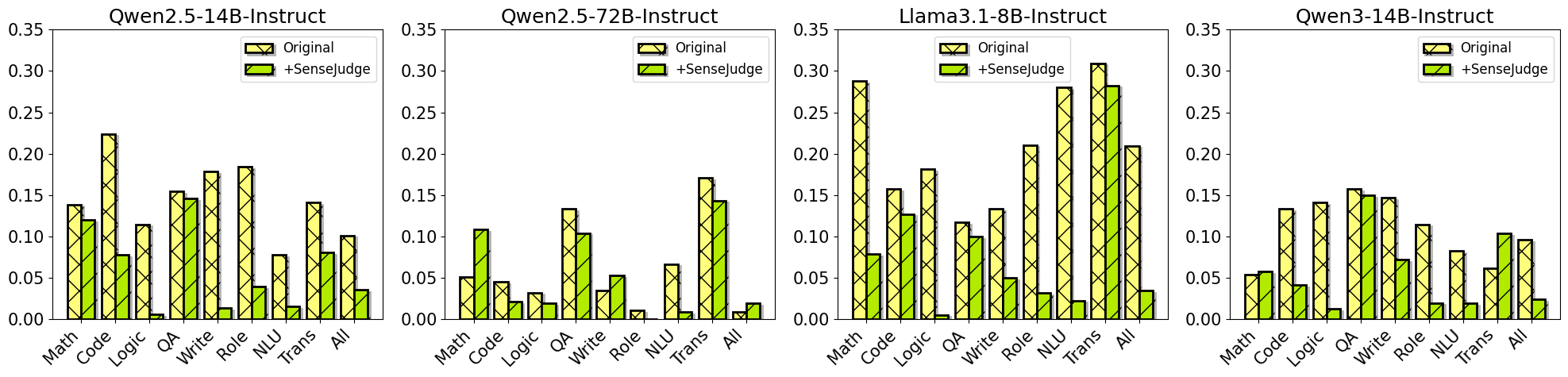}
    \caption{The \textcolor{yellow}{\rule[0.5ex]{1em}{0.8ex}} bar represents the absolute values of selecting response A (in the first place) and response B (in the second place) using the original model, while the \textcolor{mycustomgreen}{\rule[0.5ex]{1em}{0.8ex}} bar represents the absolute values after applying SenseJudge.}
    \label{fig:position}
\end{figure*}
\subsection{Main Result}

\subsubsection{LLM-as-a-Personalized-Judge}

\paragraph{Overview} In the development of business LLMs, real-time and representative human feedback is essential before the model is officially deployed. 

Many benchmarks are available to directly evaluate models' general capabilities through straightforward ground-truth answer matching.
However, this evaluation often solely considers the accuracy of model outputs, overlooking other valued aspects such as creativity and formatting, thus failing to provide a comprehensive and user-centered quality assessment.
Additionally, as many benchmark ground-truth answers may have been inadvertently or deliberately included in training data, it is difficult to assess the true performance of models using existing benchmarks alone.
Hiring a stable team to evaluate models on consistent datasets after each update is a common practice. However, as model updates become more frequent, the repetitive and labor-intensive task of human annotation is increasingly constrained by cost and efficiency. 
Therefore, ``LLM-as-a-Personalized-Judge'' has become a highly promising task that involves utilizing LLMs to simulate real user judgments of model responses, thereby streamlining and automating the model development process.

\paragraph{Comparison with Baselines}

As shown in Table \ref{tab:main}, the SenseJudge framework achieves an average Overall score of 70.66 in the LLM-as-a-Personalized-Judge task, outperforming baselines including strong open-source and closed-source models, trained judgers, and reward models by 10.68 percentage points on average.
Compared with the corresponding base models, SenseJudge yields an average gain of 7.37 percentage points across all evaluation categories and an average Overall gain of 7.38 percentage points.
Even when applied to small-scale LLMs like 8B and 14B, which are not specifically trained as judgers, it improves the Overall score by 7.41 percentage points on average.
Additionally, SenseJudge demonstrates consistent Overall improvements and broad category-level gains compared to the base models, with only minor decreases for Qwen3-14B-Instruct on Math and QA.

Although reward models such as INF-ORM-Llama3.1-70B and QRM-Gemma-2-27B have excelled on the public RewardBench leaderboard, their accuracy on the personalized dataset remains below 65\% and even falls short compared to some strong baseline LLMs. 
Autoj-13B performs even worse, achieving scores similar to random selection. 
This indicates that while judgers designed for specific judgment tasks and reward models developed for model alignment are trained on extensive preference data, they often learn fixed preferences that struggle to generalize to diverse and complex real-world scenarios and adapt to varying user preferences.

\paragraph{Positional Bias}

Previous studies~\cite{Wang2023LargeLM} have demonstrated that the relative position of two responses is an element that theoretically should be irrelevant but significantly influences the judgment result in the pairwise judgment setting.
Some LLMs exhibited a significant tendency to select responses in the first or second positions, which we refer to as \textit{position bias}.
To evaluate the impact of SenseJudge on position bias, we counted the distribution of model selection across each position under identical ground-truth conditions and calculated the absolute values of these selections. Higher absolute values indicate a stronger positional bias. As shown in Figure \ref{fig:position}, the SenseJudge framework mitigates positional bias, particularly for smaller models where this bias is often more pronounced.

\begin{figure*}[t]
    \centering
    \includegraphics[width=0.95\linewidth]{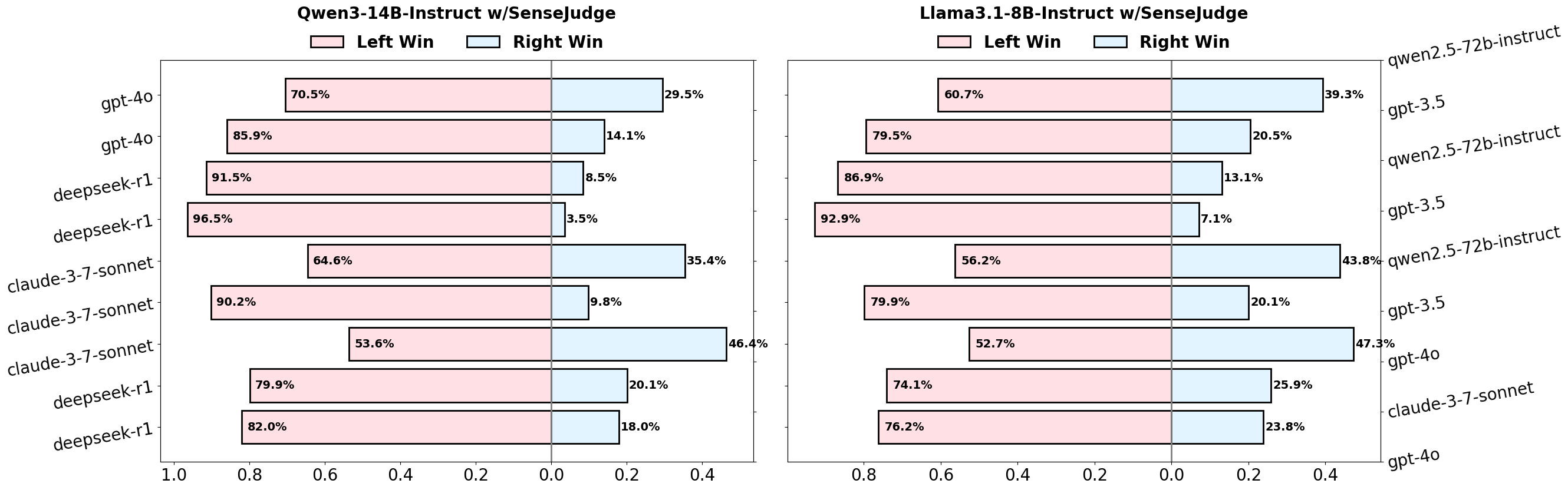}
    \caption{Pairwise judgments by Qwen3-14B-Instruct and Llama3.1-8B-Instruct with SenseJudge across advanced models (GPT-4o, DeepSeek-R1, and Claude-3-7-Sonnet), a previous-generation model (GPT-3.5), and Qwen2.5-72B-Instruct. The \textcolor{mycustompink}{\rule[0.5ex]{1em}{0.8ex}} bar indicates the win rate of the left model on the model ranking set, and the \textcolor{mycustomblue}{\rule[0.5ex]{1em}{0.8ex}} bar represents the win rate of the right model.}
    \label{fig:ranking}
\end{figure*}

\paragraph{Consistency}

\begin{table}[htbp]
    \centering
    \resizebox{\columnwidth}{!}{        \begin{tabular}{lcc}
            \toprule
            \textbf{Model} & \textbf{Original}& \textbf{SenseJudge}  \\
            \midrule
            \textbf{Qwen2.5-14B-Instruct} & 69.97 &	\textbf{74.17} \\
            \textbf{Qwen2.5-72B-Instruct} & 78.86 & 78.79 \\
            \textbf{Llama3.1-8B-Instruct } & 60.36 & \textbf{68.19} \\
            \textbf{Qwen3-14B-Instruct} & 81.23 & \textbf{81.30} \\
            \bottomrule
        \end{tabular}
    }
    \caption{Consistency performance of original models and models with SenseJudge framework.}
    \label{tab:consistency}
\end{table}

We examined the consistency of the SenseJudge framework under two distinct input settings: (query, chosen, rejected) and (query, rejected, chosen). We considered the judger's selection to be consistent if it identified the same response (either the initially designated `chosen' or `rejected') as superior across both input configurations.
Results are presented in Table \ref{tab:consistency}. Notably, the SenseJudge framework shows improvements over the base models such as Qwen2.5-14B-Instruct and Llama3.1-8B-Instruct, achieving increases in consistency scores of 4.2\% and 7.83\%, respectively, compared to their original consistency scores of 69.97\% and 60.36\%. 
While the Qwen2.5-72B-Instruct and Qwen3-14B-Instruct models exhibit high consistency in their original evaluations (78.86\% and 81.23\%), SenseJudge maintains comparable consistency.

\subsubsection{Model Ranking}

Pairwise judgment methods inherently provide a relative ranking between two model responses within a single comparison. The absolute ranking derived from such pairwise comparisons has not been thoroughly evaluated.

Therefore, based on the human-preference model rankings reported by the Arena leaderboard\footnote{\url{https://lmarena.ai/?leaderboard}}, we selected the models DeepSeek-R1, Claude-3-7-Sonnet, GPT-4o, Qwen2.5-72B-Instruct, and GPT-3.5 (ordered according to their ranking on the leaderboard) as our target models for ranking, and employed their corresponding responses from the Model Ranking Set. Among these models, DeepSeek-R1, Claude-3-7-Sonnet, and GPT-4o represent advanced models and exhibit similar performance. We then conducted pairwise comparisons using SenseJudge, randomly selecting 5 preferences from the previously generated preference set for each comparison. The results are depicted in Figure \ref{fig:ranking}.

We observed that both the relative ranking from pairwise judgment between any two models and the absolute ranking derived from aggregating all pairwise results align with the Arena human-preference ranking.
The results also demonstrate that a full all-pairs comparison is not strictly necessary to obtain a final model ranking using a pairwise approach. Taking DeepSeek-R1 as an example, its win rates when compared against all other models can be effectively used to rank all models.

\subsection{Ablation Studies}

\label{app:more_exp}
\paragraph{Performance on RewardBench}

\begin{table}[htbp]
    \centering
    \resizebox{0.45\textwidth}{!}{        \begin{tabular}{lc}
            \toprule
            \textbf{Method} & \textbf{RewardBench}  \\
            \midrule
            \textbf{SenseJudge} & 90.55 \\
            \textbf{Skywork-Critic-Llama3.1-70B} & 92.2 \\
            \textbf{Gemini-1.5-pro-0514} & 88.2 \\
            \textbf{GPT-4o-2024-05-13} & 84.6 \\
            \textbf{Meta-Llama-3.1-70B-Instruct} & 84 \\
            \textbf{Claude-3-opus-20240229} & 80.1 \\
            \bottomrule
        \end{tabular}    }
    \caption{RewardBench accuracy of different judgment methods.}
    \label{tab:rewardbench}
\end{table}

Using Qwen2.5-72B-Instruct, we applied SenseJudge to RewardBench to evaluate the general effectiveness of our approach in judgment tasks. 
RewardBench integrates several authoritative evaluation benchmarks, such as MT-Bench, AlpacaEval, and LLMBar. We selected several advanced LLM-as-a-judge methods and specifically trained generative reward models for comparison on RewardBench. For the LLM-as-judge methods, we used publicly available results from the leaderboard directly. 
Similar to the steps we followed when performing personalized judging, we sampled a subset of the data from RewardBench as a dev set for generating preferences.
As shown in Table \ref{tab:rewardbench}, SenseJudge outperformed the compared general LLM-as-a-judge methods. Without training, its gap with the generative reward model was 1.65\%.

\paragraph{Number of Dialogue Turns in the Judgment Input}

\begin{table}[htbp]
    \centering
    \resizebox{\columnwidth}{!}{        \begin{tabular}{lccc}
            \toprule
            \textbf{Number of Dialogue Turns in Input} & \textbf{(q, c, r)} & \textbf{(q, r, c)} & \textbf{Total} \\
            \midrule
            \textbf{Only Last Turn} & 58.63 & 74.51 & 66.57 \\
            \textbf{Last Two Turns} & 70.40 & 65.14 & 67.77 \\
            \textbf{All Turns within Token Limit} & 67.79 & 71.55 & \textbf{69.67} \\
            \bottomrule
        \end{tabular}    }
    \caption{Evaluation results with varying number of dialogue turns used in input. (q, c, r) denotes the input setting (query, chosen response, rejected response), and (q, r, c) denotes the input setting (query, rejected response, chosen response). The numbers represent the percentage of correct judgments.}
    \label{tab:dialogue_turns}
\end{table}

In real human dialogue scenarios, a conversation often includes several turns of historical dialogue, some relevant to the current instruction and some not. 
Therefore, determining which historical turns to consider in the input is a crucial aspect of judgment. 
To assess the impact of the number of dialogue turns used for evaluation, we experimented with using only the last turn, using the last two turns, and using all available turns within token limit in the input query. As shown in Table \ref{tab:dialogue_turns}, utilizing all available turns within the token limit achieves the best overall accuracy, although the best setting varies across input orders.

%% file: section/appd.tex
\subsection{Discussion}
\label{app:data}

\subsubsection{Details of SenseBench}
We provide the statistics of SenseBench in Table \ref{tab:sensebench_stats}.
We provide the category information of SenseBench in Table
\ref{tab:llm_evaluation_categories_wide}
We provide data examples for the category of creative writing in Table \ref{tab:creative-writing-example}.

\begin{table}[h!]
\centering
\begin{tabular}{lc}
\toprule
\textbf{Property}               & \textbf{Value}          \\
\midrule
Query Count & 400 \\
Query Average Token Count  & 956          \\
Maximum Query Token Count  & 10787          \\
Minimum Query Token Count  & 10             \\
Average Query Dialogue Turns & 4            \\
Maximum Query Dialogue Turns & 9              \\
Minimum Query Dialogue Turns & 1              \\
Personalized Subset Count & 3000 \\
Ranking Subset Count & 26400 \\
\toprule
\end{tabular}
\caption{Data Statistics of SenseBench}
\label{tab:sensebench_stats}
\end{table}

\subsubsection{Dataset Comparison}
Our data source differs from prior benchmarks such as WildChat~\cite{zhao2024wildchat1mchatgptinteraction}, LMSYS-1M~\cite{zheng2024lmsyschat1mlargescalerealworldllm}, and WildBench~\cite{lin2024wildbenchbenchmarkingllmschallenging}. SenseBench encompasses the primary task categories present in WildChat, LMSYS-1M, and WildBench. However, a defining characteristic of SenseBench is its balanced category distribution. Specifically, we curate eight core categories—Code, Translation, Role, NLU, Math, Logic, Writing, and QA—each comprising 12.5\% of the dataset. In contrast, existing datasets exhibit significant skewness. For example, WildChat is dominated by Assistance/Creative Writing (61.9\%), while other categories such as Programming (6.7\%) and Mathematical Reasoning (6.1\%) are underrepresented. Similarly, LMSYS-1M shows uneven coverage across Code (30.64\%), NLU/Translation (19.06\%), Role-playing (19.89\%), and QA (25.41\%), with overlapping annotations further complicating the distribution. WildBench also demonstrates variability, with categories such as Reasoning/Planning (23\%) and Information Seeking (17\%) being more prominent than others like Brainstorming (2\%) or Role-playing (2\%).

\subsubsection{Annotator Information}

Annotator 1 has a background in Digital Media Technology, Law, and Artificial Intelligence. Annotator 2 is trained in Materials Science and Water \& Wastewater Engineering. Annotator 3 has a background in Computer Science. Each annotator has accumulated more than five years of research experience in their respective domains, providing sufficiently diverse perspectives and distinct preference patterns. More detailed information about the annotators will be included in the public release.

\paragraph{Annotation Consistency}
We further examined the consistency of response preferences across the three annotators. The proportion of instances in which all annotators selected the same response is 49\%, indicating a relatively low level of agreement. As shown in Table~\ref{tab:annotator-agreement}, agreement patterns vary across task categories. This result suggests that preference variation across individuals with different academic and professional backgrounds can be substantial, and that the dataset captures inherently diverse and potentially conflicting preferences rather than a narrow or homogeneous distribution.

\begin{table}[htbp]
\centering
\resizebox{\columnwidth}{!}{\begin{tabular}{lccccccccc}
\toprule
 & Code & Trans & Role & NLU & Math & Logic & Write & QA & Avg \\
\midrule
Same & 35 & 50 & 75 & 66 & 58 & 59 & 63 & 85 & 491 \\
Different & 90 & 75 & 50 & 59 & 67 & 66 & 62 & 40 & 509 \\
\toprule
\end{tabular}
}
\caption{Agreement statistics across task categories.}
\label{tab:annotator-agreement}
\end{table}

\onecolumn

{\renewcommand{\arraystretch}{1.3}
\footnotesize
\begin{longtable}{|p{0.26\textwidth}|p{0.66\textwidth}|}
\caption{Categorization of SenseBench}
\label{tab:llm_evaluation_categories_wide}\\
\hline
\textbf{Category} & \textbf{Subcategory} \\
\hline
\endfirsthead

\caption[]{Categorization of SenseBench (continued)}\\
\hline
\textbf{Category} & \textbf{Subcategory} \\
\hline
\endhead

\hline
\multicolumn{2}{r}{\textit{Continued on next page}}\\
\endfoot

\hline
\endlastfoot

\textbf{Math} & \textbf{Elementary Mathematics}: Calculation problems, simple word problems \\
                   & \textbf{Applied Mathematics}: Word problems involving real-world scenarios \\
                   & \textbf{Advanced Mathematics}: Concept explanation, theorem proving, etc. \\
\hline
\textbf{Reasoning} & \textbf{Brain Teasers}: Various interesting brain teaser questions \\
                           & \textbf{General Logic Problems}: Problems containing basic life logic, similar to but not exactly the same as brain teasers \\
                           & \textbf{Mathematical Logical Reasoning}: Problems requiring both logical reasoning and mathematical calculation \\
                           & \textbf{Inductive Reasoning}: Pattern finding and other similar problems \\
                           & \textbf{Conditional Reasoning}: Reasoning based on multiple given conditions; sometimes requiring traversal of all possible cases \\
\hline
\textbf{Code} & \textbf{Code Generation}: User provides task requirements, model generates corresponding code (Natural Language $\rightarrow$ Code) \\
              & \textbf{Code Error Reporting}: User provides code error information, model gives possible solutions or perspectives \\
              & \textbf{Code Explanation}: User provides code, model provides explanation/comments, etc. (Code $\rightarrow$ Natural Language) \\
\hline
\textbf{Comprehensive Q\&A} & \textbf{Opinion Expression}: User outputs an opinion or asks the model for its view on something \\
                            & \textbf{Seeking Advice}: User asks the model for advice on future actions \\
                            & \textbf{General Knowledge}: User asks the model common knowledge questions, hoping for comprehensive and insightful answers \\
                            & \textbf{Other Questions}: User wants the model to generate questions in a specific role (e.g., for an interview) \\
\hline
\textbf{Writing} & \textbf{Practical Writing}: Letters, emails; speeches, applications, reflections; contracts, termination letters, bids, proposals, plans, wills \\
                          & \textbf{Creative Writing}: Writing on a specified topic, continuation writing; text expansion, imitation writing; reading notes, movie reviews, reflections; social media posts (WeChat Moments, Xiaohongshu, Weibo, etc.); listing names, nicknames, brand names, store names, dish names, etc. \\
                          & \textbf{Professional Writing}: Thesis-like (outlines, titles, abstracts, content, bibliography generation); reports (legal documents, financial analysis); solutions; explanatory texts (recipes, instructions); +Complex instruction following (Content, Situation, Style, Format) \\
\hline
\textbf{Role-Playing} & \textbf{Celebrity}: Playing the role of real celebrities (politicians, athletes, etc.) \\
                      & \textbf{Entertainment}: Games involving the user and the LLM, requiring imagination and creativity \\
                      & \textbf{Practical}: Emotional support (friends, relatives, pets, etc.); critical thinking, providing advice; accurately executing natural language tasks (proofreading, generating training data); creativity and adaptability (imitating details in real-life scenarios) \\
\hline
\textbf{Natural Language Understanding} & \textbf{Reading Comprehension}: Understanding a passage and providing answers, including intent recognition \\
                                        & \textbf{Translation Related}: Requiring the model to master different languages and understand their interactions \\
                                        & \textbf{Text Classification}: Testing the model's ability to classify text data into given categories with different characteristics \\
                                        & \textbf{Information Extraction}: Measuring the model's ability to extract knowledge from text data \\
                                        & \textbf{Text Polishing \& Editing}: Simplification, expansion, keyword extraction, error checking, and correction of grammatical and logical errors \\
\hline
\textbf{Translation} & \textbf{Basic Translation}: Accurate translation of specified text. \\
                     & \textbf{Complex Instruction Following}: Following complex instructions such as using a specific style or terminology. \\
                     & \textbf{Multi-Task Translation}: Requiring the model to perform multiple tasks simultaneously (e.g., translate and format, translate and proofread, translate + creative writing). \\
                     & \textbf{Self-Referential Translation}: Asking the model to translate its previous responses into another language to test self-generated text translation and short-term memory. \\
                     & \textbf{Long-Term Translation Memory}: Requiring the model to act as a translator long-term to test long-term memory. \\
                     & \textbf{Dialect Translation}: Ability to understand and translate dialects into standard languages, and mastery of regional linguistic features. \\
\hline
\end{longtable}
}

\begin{longtable}{|p{2.5cm}|p{12cm}|}
\caption{Creative writing example: character and weapon design in a fictional setting (Void Empire universe)}\label{tab:creative-writing-example}\\
\hline
\textbf{From} & \textbf{Content} \\
\hline
\endfirsthead

\hline
\textbf{From} & \textbf{Content} \\
\hline
\endhead

Human &
请设计以下人物:虚空帝国女武神骑士团团长——“女武神”，设计内容包括:人物姓名、外在形象、接受虚空帝国生物技术改造获得了心灵感应的超自然能力、虚空国防大学教育经历及学历学位、中将军衔、个人专长等 \\
\hline

Assistant &
\textbf{人物姓名：} 莉安娜·艾斯瑞尔 (Liana Esriel)

\textbf{外在形象：} 莉安娜拥有一头银白色的长发，通常以高级军官的严肃发型束起，突显出她冷静而坚定的眼神。她的瞳色是不寻常的深紫色，由于生物技术改造，她的眼神中仿佛能洞察人心。她的身材高挑，肌肉线条明显，显示出卓越的体能和战斗能力。她通常穿着虚空帝国女武神骑士团的标准装甲，这种装甲在提供极高防御力的同时，还特别强调了灵活性和速度，且以黑紫色为主调，显得既神秘又威严。

\textbf{超自然能力：} 通过虚空帝国生物技术改造获得心灵感应能力，可感知情绪与意图，并进行低强度思想交流。

\textbf{教育经历：} 虚空国防大学：心理战术与群体操控方向  
学位：战略学博士（论文：《心理战术在现代战争中的应用》）

\textbf{军衔：} 中将

\textbf{个人专长：} 战术规划、心理战术、指挥统合、人际领导能力 \\
\hline

Human &
请设计莉安娜用作武器和指挥剑的佩剑:“月光骑士” \\
\hline

Assistant &
\textbf{佩剑名称：} 月光骑士 (Moonlight Knight)

\textbf{外观设计：} 剑长约1.2米，虚空金属锻造，月光下呈银蓝色光泽；剑柄镶嵌黑紫宝石并刻有符文。

\textbf{功能特性：}
心灵共鸣增强直觉与预判能力；  
能量聚焦可释放高穿透斩击；  
指挥强化提升军队士气与指令传导效率。

\textbf{象征意义：} 女武神骑士团权威象征，代表统御与战场支配力。 \\
\hline

Human &
请设计莉安娜的佩枪:“心灵冲击” \\
\hline
\end{longtable}

\subsection{Prompt}
\label{app:prompt}

\begin{tcolorbox}[title = {Preference Construction Prompt}, breakable]

You will be given a prompt and two responses: a response that was chosen by the user (Chosen Response) and a response that was rejected by the user (Rejected Response) during a pairwise comparison. The prompt is a "Human" utterance containing a request or question and the responses are "Assistant" utterances that provide answers or responses for the human. Your task is to generate a concise, specific, description of the user's persona preference in about three single sentences, i.e. a persona preference. The persona preference should contain reasoning for why the user preferred and picked the Chosen Response and did not pick the Rejected Response. The persona preference should discuss higher-level preference that can be inferred about the user's persona. The persona preference should be concise and should not mention specific details or exact words and phrasing present in the prompt or responses. Answer in English.\\
\\
Query: \{query\}\\
---\\
Chosen Response: Assistant: \{chosen response\}\\
---\\
Rejected Response: Assistant:  \{rejected response\}\\
---\\
Persona Preference: \\
\end{tcolorbox}

\begin{tcolorbox}[title = {Judgment Prompt}, breakable]
You are going to evaluate two responses to a given user query and determine which response is superior. Below is the relevant content:\\

BEGIN DATA\\
***\\
User Query:\{query\}\\
***\\
Response A: \{response 1\}\\
***\\
Response B: \{response 2\}\\
***\\
END DATA\\
\\
Here are the guidelines for evaluating and comparing the two responses:\\
\\
***BEGIN User Preferences***\\
\{preference\}\\
***END User Preferences***\\
\\
1. Score each of the two responses based on the user preferences.\\
2. Based on the scores obtained in the first step, determine which response is better. If Response A is better, output ``The final decision is Response A.'' If Response B is better, output ``The final decision is Response B.''\\
\end{tcolorbox}

\subsection{Preference Case}
\label{app:preference}
\begin{tcolorbox}[title = {Preference for math tasks extracted from the development set}, breakable]

    "Based on the comparison, the user's persona demonstrates these characteristics:\par 1. They strongly value methodical reasoning that transparently explores multiple approaches and validates failures, prioritizing thorough cognitive processes over conventional solutions. \par 2. They prefer responses that explicitly build verification frameworks and test edge cases, rejecting shortcuts that lack demonstrated iterative refinement. \par 3. They seek pedagogical clarity through structured decomposition of assumptions, showing aversion to answers that prioritize memorized conclusions over original analytical scaffolding. \par This persona prefers the Chosen Response for its stepwise validation of failed strategies and truth-table proofs, while rejecting the alternative for its faster-to-conclusion approach despite equivalent correctness.",\\
    "Based on the comparison, the user's persona preference indicates they value responses that demonstrate:\par 1.  **Comprehensive Logical Reasoning:** They prefer answers that break down the scenario step-by-step, exploring potential starting points and logical implications, rather than stating a direct conclusion without thorough justification.\par 2.  **Acknowledgement of Edge Cases \& Nuance:** They appreciate responses that explicitly consider edge cases (like being the first place initially) and contextual factors, showing awareness that real-world questions often have layers beyond the surface.\par 3.  **Structured and Explicit Answer Presentation:** They favor responses that clearly summarize the primary conclusion after presenting the reasoning, making the final answer distinct and easy to identify, rather than leaving it embedded within the explanation. They reject responses perceived as overly simplistic or lacking in explanatory depth.",\\
    "Based on this pairwise comparison, the user's persona indicates a preference for solutions that prioritize structural clarity and efficient presentation over exhaustive validation. They value responses that strategically organize key steps using visual hierarchy (like bold headers and bullet points) to facilitate rapid comprehension of the core logic. The rejection of the extended verification approach suggests they prioritize concise, solution-focused explanations and consider secondary validations redundant when core reasoning is robustly established.",\\
    "Based on the user's preference, they exhibit a persona that: \par They prioritize responses that directly focus on fundamental principles and theoretical foundations over exhaustive validations or trial-and-error explorations. \par They value concise analytical reasoning that efficiently leverages well-established concepts (like binary representation) to solve complex problems. \par They implicitly expect solutions to avoid redundancy and trust core mathematical insights rather than requiring step-by-step demonstrations of every outcome.",\\
    "The user prefers clear, step-by-step logical reasoning that methodically eliminates possibilities without unnecessary complexity. They value responses that demonstrate a direct path to the solution using the given constraints, avoiding speculative or redundant tangents. This indicates a persona that prioritizes structured problem-solving and appreciates concise, easily traceable explanations grounded in deductive clarity.",\\
    "Based on the chosen solution and rejected alternative, this user strongly prefers concise explanations with efficient problem-solving frameworks over exhaustive explorations of edge cases. They prioritize clear, step-by-step methodologies that deliver optimal solutions directly, rather than approaches that delve into unnecessary hypothetical scenarios that complicate the core logic. The rejection of the longer analysis reveals an aversion to overly theoretical or redundant validations that don't tangibly improve the final answer's practicality or simplicity.",\\
    "Based on the chosen response preference, the user appears to value analytical depth over simplistic explanations. They likely seek assistants who systematically unpack layered logical contradictions rather than settling for surface-level fixes. This user probably prefers intellectual rigor in problem-solving, where potential ambiguities are thoroughly examined through contextual, mathematical, and semantic lenses. They seem to appreciate responses that acknowledge plausible interpretations before resolving them.",\\
    "Based on the preference for the Chosen Response over the Rejected Response, the user's persona demonstrates:\par 1. They value analytical rigor and systematic problem-solving approaches, seeking responses that methodically break down constraints and explore multiple strategies rather than presenting isolated solutions without justification. \par 2. They prefer responses that optimize for efficiency by testing different scenarios and validating the optimal solution, rejecting approaches that overlook practical time-saving tactics or introduce unnecessary steps. \par 3. Their learning style prioritizes conceptual clarity over fragmented execution, favoring explanations that emphasize logical reasoning patterns applicable to similar challenges rather than ad-hoc step sequences.",\\
    "Based on the preferred response, the user values thorough, methodical explanations that explicitly outline academic reasoning processes, including problem decomposition and solution verification, over concise solutions that skip foundational steps. This suggests the user prioritizes pedagogical clarity and structured learning for full conceptual understanding, likely indicating an educational context or self-learning scenario where process is emphasized. The rejection of the simpler calculation implies the user seeks responses that model systematic thinking rather than just providing correct answers.",\\
    "The user prefers responses that demonstrate precise, mathematically sound reasoning while avoiding unnecessary complexity. They value concise explanations that distill core strategic principles into actionable insights without overcomplicating the analysis. This persona favors logical depth presented efficiently, rejecting responses with convoluted deductions even if equally thorough, indicating a prioritization of clarity and applicability in problem-solving."
\end{tcolorbox}

\begin{tcolorbox}[title = {Filtered preference set obtained after selection}, breakable]

    "Based on this pairwise comparison, the user's persona indicates a preference for solutions that prioritize structural clarity and efficient presentation over exhaustive validation. They value responses that strategically organize key steps using visual hierarchy (like bold headers and bullet points) to facilitate rapid comprehension of the core logic. The rejection of the extended verification approach suggests they prioritize concise, solution-focused explanations and consider secondary validations redundant when core reasoning is robustly established.",\\

    "The user prefers responses that demonstrate precise, mathematically sound reasoning while avoiding unnecessary complexity. They value concise explanations that distill core strategic principles into actionable insights without overcomplicating the analysis. This persona favors logical depth presented efficiently, rejecting responses with convoluted deductions even if equally thorough, indicating a prioritization of clarity and applicability in problem-solving."
\end{tcolorbox}